\documentclass[10pt,twocolumn,letterpaper]{article}

\usepackage[pagenumbers]{cvpr} 

\usepackage[dvipsnames]{xcolor}

\usepackage{amsmath}
\usepackage{amssymb}
\usepackage{times}
\usepackage{bm}
\usepackage{xargs}
\usepackage{multirow}
\usepackage{amsfonts}
\usepackage{color}
\usepackage{booktabs}

\usepackage{algorithm}
\usepackage{algorithmic}
\usepackage{graphicx}
\usepackage{subfiles}
\usepackage{enumitem}
\usepackage{wrapfig}
\usepackage{capt-of}

\definecolor{cvprblue}{rgb}{0.21,0.49,0.74}
\usepackage[breaklinks,colorlinks,citecolor=cvprblue]{hyperref}

\def\paperID{****} 
\def\confName{CVPR}
\def\confYear{2024}

\title{Factorized Diffusion Architectures for\\Unsupervised Image Generation and Segmentation}

\author{Xin Yuan\\%
University of Chicago\\%
{\tt\small{yuanx@uchicago.edu}}%
\and
Michael Maire\\%
University of Chicago\\%
{\tt\small{mmaire@uchicago.edu}}}

\begin{document}


\newcommand{\labelFig}[1]{\label{fig:#1}}
\newcommand{\refFig}[1]{Figure~\ref{fig:#1}}
\newcommand{\labelEqn}[1]{\label{eqn:#1}}
\newcommand{\refEqn}[1]{(\ref{eqn:#1})}
\newcommand{\labelSec}[1]{\label{sec:#1}}
\newcommand{\refSec}[1]{Section \ref{sec:#1}}
\newcommand{\labelAlg}[1]{\label{alg:#1}}
\newcommand{\refAlg}[1]{Alg.~\ref{alg:#1}}
\newcommand{\labelTab}[1]{\label{tab:#1}}
\newcommand{\refTab}[1]{Table~\ref{tab:#1}}

%
%
%
%

\newcommand{\ba}{{\bm{a}}}
\newcommand{\bb}{{\bm{b}}}
\newcommand{\bc}{{\bm{c}}}
\newcommand{\bd}{{\bm{d}}}
\newcommand{\be}{{\bm{e}}}
\newcommand{\bg}{{\bm{g}}}
\newcommand{\bh}{{\bm{h}}}
\newcommand{\bi}{{\bm{i}}}
\newcommand{\bj}{{\bm{j}}}
\newcommand{\bk}{{\bm{k}}}
\newcommand{\bl}{{\bm{l}}}
\newcommand{\bn}{{\bm{n}}}
\newcommand{\bo}{{\bm{o}}}
\newcommand{\bp}{{\bm{p}}}
\newcommand{\bq}{{\bm{q}}}
\newcommand{\br}{{\bm{r}}}
\newcommand{\bs}{{\bm{s}}}
\newcommand{\bt}{{\bm{t}}}
\newcommand{\bu}{{\bm{u}}}
\newcommand{\bv}{{\bm{v}}}
\newcommand{\bw}{{\bm{w}}}
\newcommand{\bx}{{\bm{x}}}
\newcommand{\by}{{\bm{y}}}
\newcommand{\bz}{{\bm{z}}}

\newcommand{\bA}{{\bm{A}}}
\newcommand{\bB}{{\bm{B}}}
\newcommand{\bC}{{\bm{C}}}
\newcommand{\bD}{{\bm{D}}}
\newcommand{\bE}{{\bm{E}}}
\newcommand{\bF}{{\bm{F}}}
\newcommand{\bG}{{\bm{G}}}
\newcommand{\bH}{{\bm{H}}}
\newcommand{\bI}{{\bm{I}}}
\newcommand{\bJ}{{\bm{J}}}
\newcommand{\bK}{{\bm{K}}}
\newcommand{\bL}{{\bm{L}}}
\newcommand{\bM}{{\bm{M}}}
\newcommand{\bN}{{\bm{N}}}
\newcommand{\bO}{{\bm{O}}}
\newcommand{\bP}{{\bm{P}}}
\newcommand{\bQ}{{\bm{Q}}}
\newcommand{\bR}{{\bm{R}}}
\newcommand{\bS}{{\bm{S}}}
\newcommand{\bT}{{\bm{T}}}
\newcommand{\bU}{{\bm{U}}}
\newcommand{\bV}{{\bm{V}}}
\newcommand{\bW}{{\bm{W}}}
\newcommand{\bX}{{\bm{X}}}
\newcommand{\bY}{{\bm{Y}}}
\newcommand{\bZ}{{\bm{Z}}}

\newcommand{\ea}{{\emph{a}}}
\newcommand{\eb}{{\emph{b}}}
\newcommand{\ec}{{\emph{c}}}
\newcommand{\ed}{{\emph{d}}}
\newcommand{\ee}{{\emph{e}}}
\newcommand{\ef}{{\emph{f}}}
\newcommand{\eh}{{\emph{h}}}
\newcommand{\ei}{{\emph{i}}}
\newcommand{\ej}{{\emph{j}}}
\newcommand{\ek}{{\emph{k}}}
\newcommand{\el}{{\emph{l}}}
\newcommand{\en}{{\emph{n}}}
\newcommand{\eo}{{\emph{o}}}
\newcommand{\ep}{{\emph{p}}}
\newcommand{\eq}{{\emph{q}}}
\newcommand{\er}{{\emph{r}}}
\newcommand{\es}{{\emph{s}}}
\newcommand{\et}{{\emph{t}}}
\newcommand{\eu}{{\emph{u}}}
\newcommand{\ev}{{\emph{v}}}
\newcommand{\ew}{{\emph{w}}}
\newcommand{\ex}{{\emph{x}}}
\newcommand{\ey}{{\emph{y}}}
\newcommand{\ez}{{\emph{z}}}

\newcommand{\eA}{{\emph{A}}}
\newcommand{\eB}{{\emph{B}}}
\newcommand{\eC}{{\emph{C}}}
\newcommand{\eD}{{\emph{D}}}
\newcommand{\eE}{{\emph{E}}}
\newcommand{\eF}{{\emph{F}}}
\newcommand{\eG}{{\emph{G}}}
\newcommand{\eH}{{\emph{H}}}
\newcommand{\eI}{{\emph{I}}}
\newcommand{\eJ}{{\emph{J}}}
\newcommand{\eK}{{\emph{K}}}
\newcommand{\eL}{{\emph{L}}}
\newcommand{\eM}{{\emph{M}}}
\newcommand{\eN}{{\emph{N}}}
\newcommand{\eO}{{\emph{O}}}
\newcommand{\eP}{{\emph{P}}}
\newcommand{\eQ}{{\emph{Q}}}
\newcommand{\eR}{{\emph{R}}}
\newcommand{\eS}{{\emph{S}}}
\newcommand{\eT}{{\emph{T}}}
\newcommand{\eU}{{\emph{U}}}
\newcommand{\eV}{{\emph{V}}}
\newcommand{\eW}{{\emph{W}}}
\newcommand{\eX}{{\emph{X}}}
\newcommand{\eY}{{\emph{Y}}}
\newcommand{\eZ}{{\emph{Z}}}

\newcommand{\ta}{{\textbf{a}}}
\newcommand{\tb}{{\textbf{b}}}
\newcommand{\tc}{{\textbf{c}}}
\newcommand{\td}{{\textbf{d}}}
\newcommand{\te}{{\textbf{e}}}
\newcommand{\tf}{{\textbf{f}}}
\newcommand{\tg}{{\textbf{g}}}
\newcommand{\ti}{{\textbf{i}}}
\newcommand{\tj}{{\textbf{j}}}
\newcommand{\tk}{{\textbf{k}}}
\newcommand{\tl}{{\textbf{l}}}
\newcommand{\tm}{{\textbf{m}}}
\newcommand{\tn}{{\textbf{n}}}
\newcommand{\tp}{{\textbf{p}}}
\newcommand{\tq}{{\textbf{q}}}
\newcommand{\tr}{{\textbf{r}}}
\newcommand{\tu}{{\textbf{u}}}
\newcommand{\tv}{{\textbf{v}}}
\newcommand{\tw}{{\textbf{w}}}
\newcommand{\tx}{{\textbf{x}}}
\newcommand{\ty}{{\textbf{y}}}
\newcommand{\tz}{{\textbf{z}}}

\newcommand{\tA}{{\textbf{A}}}
\newcommand{\tB}{{\textbf{B}}}
\newcommand{\tC}{{\textbf{C}}}
\newcommand{\tD}{{\textbf{D}}}
\newcommand{\tE}{{\textbf{E}}}
\newcommand{\tF}{{\textbf{F}}}
\newcommand{\tG}{{\textbf{G}}}
\newcommand{\tH}{{\textbf{H}}}
\newcommand{\tI}{{\textbf{I}}}
\newcommand{\tJ}{{\textbf{J}}}
\newcommand{\tK}{{\textbf{K}}}
\newcommand{\tL}{{\textbf{L}}}
\newcommand{\tM}{{\textbf{M}}}
\newcommand{\tN}{{\textbf{N}}}
\newcommand{\tO}{{\textbf{O}}}
\newcommand{\tP}{{\textbf{P}}}
\newcommand{\tQ}{{\textbf{Q}}}
\newcommand{\tR}{{\textbf{R}}}
\newcommand{\tS}{{\textbf{S}}}
\newcommand{\tT}{{\textbf{T}}}
\newcommand{\tU}{{\textbf{U}}}
\newcommand{\tV}{{\textbf{V}}}
\newcommand{\tW}{{\textbf{W}}}
\newcommand{\tX}{{\textbf{X}}}
\newcommand{\tY}{{\textbf{Y}}}
\newcommand{\tZ}{{\textbf{Z}}}

\newcommand{\mA}{{\mathcal{A}}}
\newcommand{\mB}{{\mathcal{B}}}
\newcommand{\mC}{{\mathcal{C}}}
\newcommand{\mD}{{\mathcal{D}}}
\newcommand{\mE}{{\mathcal{E}}}
\newcommand{\mF}{{\mathcal{F}}}
\newcommand{\mG}{{\mathcal{G}}}
\newcommand{\mH}{{\mathcal{H}}}
\newcommand{\mI}{{\mathcal{I}}}
\newcommand{\mJ}{{\mathcal{J}}}
\newcommand{\mK}{{\mathcal{K}}}
\newcommand{\mL}{{\mathcal{L}}}
\newcommand{\mM}{{\mathcal{M}}}
\newcommand{\mN}{{\mathcal{N}}}
\newcommand{\mO}{{\mathcal{O}}}
\newcommand{\mP}{{\mathcal{P}}}
\newcommand{\mQ}{{\mathcal{Q}}}
\newcommand{\mR}{{\mathcal{R}}}
\newcommand{\mS}{{\mathcal{S}}}
\newcommand{\mT}{{\mathcal{T}}}
\newcommand{\mU}{{\mathcal{U}}}
\newcommand{\mV}{{\mathcal{V}}}
\newcommand{\mW}{{\mathcal{W}}}
\newcommand{\mX}{{\mathcal{X}}}
\newcommand{\mY}{{\mathcal{Y}}}
\newcommand{\mZ}{{\mathcal{Z}}}

\newcommand{\norm}[1]{\left\lVert #1 \right\rVert}
\newcommand{\degA}[1]{#1\textdegree{}}
\newcommand{\PreserveBackslash}[1]{\let\temp=\\#1\let\\=\temp}

\def\eg{\emph{e.g.,}} \def\Eg{\emph{E.g.,}}          
\def\ie{\emph{i.e.,}} \def\Ie{\emph{I.e.,}}          
\def\cf{\emph{c.f.}} \def\Cf{\emph{C.f.}}          
\def\etc{\emph{etc.}}                              
\def\vs{\emph{vs.}}                                
\def\etal{\emph{et al.}}                           

\maketitle
\begin{abstract}
We develop a neural network architecture which, trained in an unsupervised manner as a denoising diffusion model, simultaneously learns to both generate and segment images.  Learning is driven entirely by the denoising diffusion objective, without any annotation or prior knowledge about regions during training.  A computational bottleneck, built into the neural architecture, encourages the denoising network to partition an input into regions, denoise them in parallel, and combine the results.  Our trained model generates both synthetic images and, by simple examination of its internal predicted partitions, a semantic segmentation of those images.  Without any finetuning, we directly apply our unsupervised model to the downstream task of segmenting real images via noising and subsequently denoising them.  Experiments demonstrate that our model achieves accurate unsupervised image segmentation and high-quality synthetic image generation across multiple datasets.

\end{abstract}

\section{Introduction}
\label{sec:intro}
\begin{figure*}[tbh]
   \begin{center}
   \begin{minipage}[t]{0.49\linewidth}
      \vspace{0pt}
      \centering
      \subfloat[\textbf{\textsf{\scriptsize{Simultaneous Image and Region Generation}}}]{
        \includegraphics[width=0.95\linewidth]{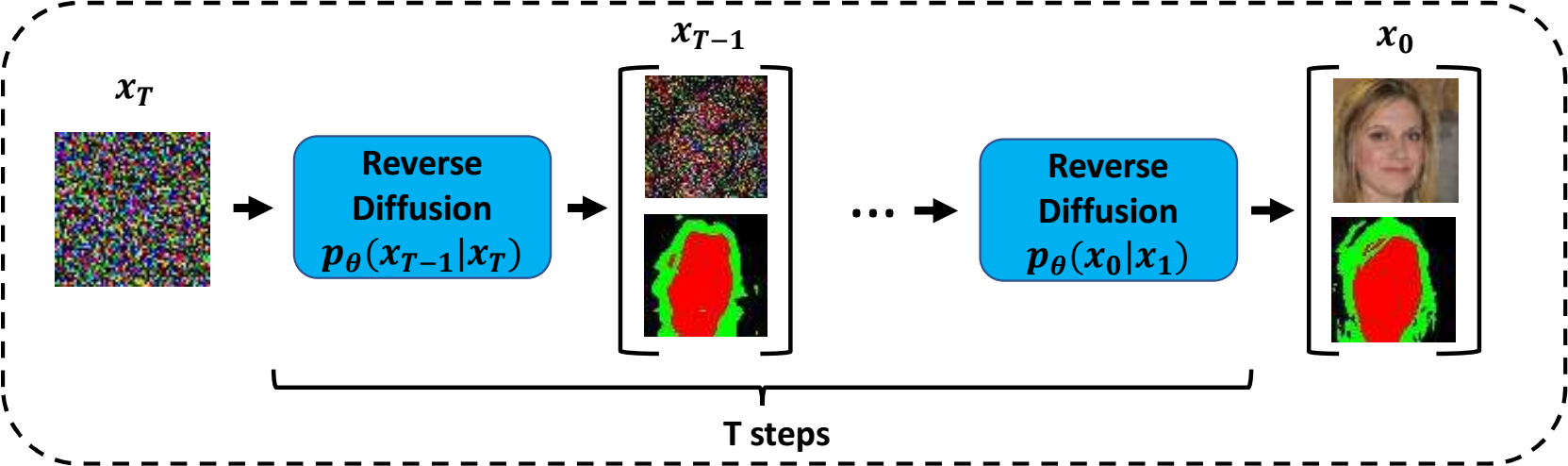}
      }
   \end{minipage}%
   \hfill%
   \begin{minipage}[t]{0.49\linewidth}
      \vspace{0pt}
      \centering
      \subfloat[\textbf{\textsf{\scriptsize{Segmentation of a Novel Input Image}}}]{
         \includegraphics[width=0.95\linewidth]{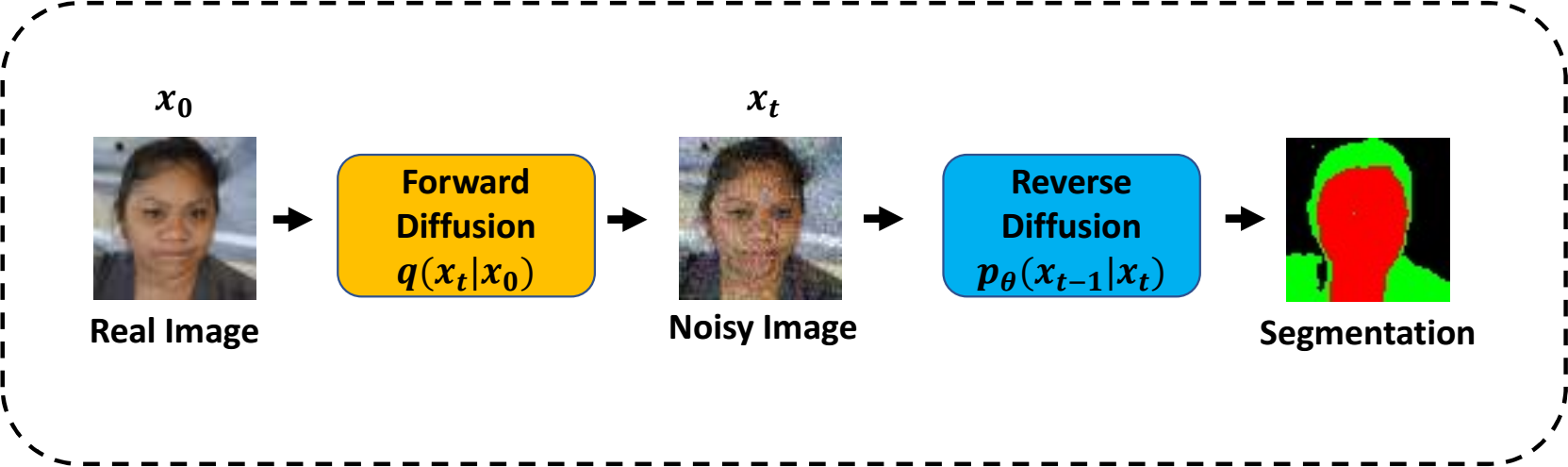}
      }
   \end{minipage}
   \end{center}%
   \vspace{-15pt}%
   \begin{center}%
   \begin{minipage}[t]{0.49\linewidth}
      \vspace{0pt}
      \centering
      \subfloat[\textbf{\textsf{\scriptsize{Generated Images}}}] {
         \includegraphics[width=0.45\linewidth]{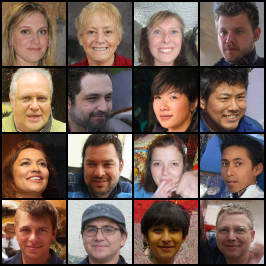}
      }
      \subfloat[\textbf{\textsf{\scriptsize{Generated Regions}}}] {
         \includegraphics[width=0.45\linewidth]{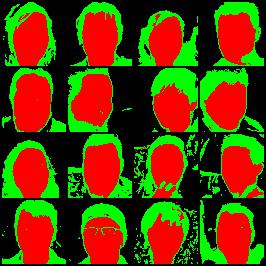}
      }
   \end{minipage}%
   \hfill%
   \begin{minipage}[t]{0.49\linewidth}
      \vspace{0pt}
      \centering
      \subfloat[\textbf{\textsf{\scriptsize{Real Images}}}] {
         \includegraphics[width=0.45\linewidth]{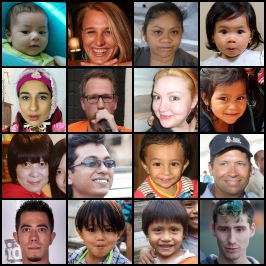}
      }
      \subfloat[\textbf{\textsf{\scriptsize{Segmentations}}}] {
         \includegraphics[width=0.45\linewidth]{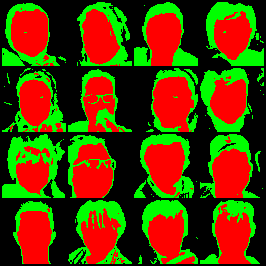}
      }
   \end{minipage}
   \end{center}
   \vspace{-15pt}
   \caption{%
      \textbf{Unifying image generation and segmentation.}
      (a) We design a denoising diffusion model with a specific architecture that couples region prediction with
      spatially-masked diffusion over predicted regions, thereby generating both simultaneously.
      (b) An additional byproduct of running our trained denoising model on an arbitrary input image is a segmentation
      of that image.
      Using a model trained on FFHQ~\cite{DBLP:conf/cvpr/KarrasLA19}, we achieve both high quality synthesis of images
      and corresponding semantic segmentations (c-d), as well as the ability to accurately segment images of real
      faces (e-f).  Segmenting a real image is fast, requiring only one forward pass (one denoising step).%
   }%
   \label{fig:overview}
\end{figure*}

Supervised deep learning yields powerful discriminative representations, and has fundamentally advanced many computer
vision tasks, including
image classification~\cite{
DBLP:conf/cvpr/DengDSLL009,
DBLP:journals/corr/SimonyanZ14a,
DBLP:conf/cvpr/HeZRS16,
DBLP:conf/cvpr/HuangLMW17},
object detection~\cite{
DBLP:conf/cvpr/GirshickDDM14,
DBLP:conf/cvpr/RedmonDGF16,
DBLP:conf/eccv/LiuAESRFB16},
and semantic and instance segmentation~\cite{
DBLP:conf/cvpr/LongSD15,
DBLP:conf/iccv/HeGDG17,
kirillov2023segany}.
Yet, annotation efforts~\cite{DBLP:conf/cvpr/DengDSLL009}, especially those involving fine-grained labeling for tasks
such as segmentation~\cite{DBLP:conf/eccv/LinMBHPRDZ14}, can become prohibitively expensive to scale with increasing
dataset size.  This motivates an ongoing revolution in self-supervised methods for visual representation learning,
which do not require any annotated data during a large-scale pre-training phase~\cite{
DBLP:conf/iccv/CaronBMJ19,
DBLP:conf/iccv/DoerschGE15,
DBLP:conf/eccv/ZhangIE16,
LMS:CVPR:2017,
DBLP:conf/cvpr/He0WXG20,
DBLP:journals/corr/abs-2002-05709,
DBLP:journals/corr/abs-2003-04297}.
However, many of these approaches, including those in the particularly successful contrastive learning paradigm~\cite{
DBLP:conf/cvpr/He0WXG20,
DBLP:journals/corr/abs-2002-05709,
DBLP:journals/corr/abs-2003-04297},
still require supervised fine-tuning (\emph{e.g.,} linear probing) on labeled data to adapt networks
to downstream tasks such as classification~\cite{DBLP:conf/cvpr/He0WXG20,DBLP:journals/corr/abs-2002-05709}
or segmentation~\cite{caron2021emerging, DBLP:conf/nips/ZhangM20}.

In parallel with the development of self-supervised deep learning, rapid progress on a variety of frameworks for
deep generative models~\cite{
DBLP:journals/corr/KingmaW13,
goodfellow2014generative,
Xu18,
Han17,
Oord16,
DBLP:journals/corr/abs-1809-09087,
DBLP:conf/nips/HoJA20,
DBLP:conf/iclr/SongME21,
DBLP:conf/cvpr/RombachBLEO22}
has lead to new systems for high-quality image synthesis.  This progress inspires efforts to explore representation
learning within generative models, with recent results suggesting that image generation can serve as a good proxy
task for capturing high-level semantic information, while also enabling realistic image synthesis.

Building upon generative adversarial networks (GANs)~\cite{goodfellow2014generative} or variational autoencoders
(VAEs)~\cite{DBLP:journals/corr/KingmaW13}, InfoGAN~\cite{DBLP:conf/nips/ChenCDHSSA16} and Deep
InfoMax~\cite{DBLP:conf/iclr/HjelmFLGBTB19} demonstrate that generative models can perform image classification
without any supervision.  PerturbGAN~\cite{DBLP:conf/nips/BielskiF19} focuses on a more complex task, unsupervised
image segmentation, by forcing an encoder to map an image to the input of a pre-trained generator so that it
synthesizes a composite image that matches the original input image.  However, here training is conducted in
two stages and mask generation relies on knowledge of predefined object classes.

Denoising diffusion probabilistic models (DDPMs)~\cite{DBLP:conf/nips/HoJA20} also achieve impressive performance in
generating realistic images.  DatasetDDPM~\cite{DBLP:conf/iclr/BaranchukVRKB22} investigates the intermediate
activations from the pre-trained U-Net~\cite{DBLP:conf/miccai/RonnebergerFB15} network that approximates the Markov
step of the reverse diffusion process in DDPM, and proposes a simple semantic segmentation pipeline fine-tuned on a few
labeled images.  In spite of this usage of labels, DatasetDDPM demonstrates that high-level semantic information,
which is valuable for downstream vision tasks, can be extracted from pre-trained DDPM U-Net.
Diff-AE~\cite{DBLP:conf/cvpr/PreechakulCWS22} and PADE~\cite{DBLP:conf/nips/ZhangZL22} are recently proposed methods
for representation learning by reconstructing images in the DDPM framework.  However, their learned representations
are in the form of a latent vector containing information applicable for image classification.

In contrast to all of these methods, we demonstrate a fundamentally new paradigm for unsupervised visual
representation learning with generative models: constrain the architecture of the model with a structured
bottleneck that provides an interpretable view of the generation process, and from which one can simply read off
desired latent information.  This structured bottleneck does not exist in isolation, but rather is co-designed
alongside the network architecture preceding and following it.  The computational layout of these pieces must
work together in a manner that forces the network, when trained from scratch for generation alone, to populate the
bottleneck data structure with an interpretable visual representation.

We demonstrate this concept in the scenario of a DDPM for image generation and the selection of semantic
segmentation as the interpretable representation to be read from the bottleneck.  Thus, we frame unsupervised image
segmentation and generation in a unified system.  Moreover, experiments demonstrate that domain-specific
bottleneck design not only allows us to accomplish an end task (segmentation) for free, but also boosts the
quality of generated samples.  This challenges the assumption that generic architectures (\emph{e.g.,}
Transformers) alone suffice; we find synergy by organizing such generic building blocks into a factorized
architecture which generates different image regions in parallel.

\begin{figure*}[h]
   \begin{center}
      \centerline{\includegraphics[width=\linewidth]{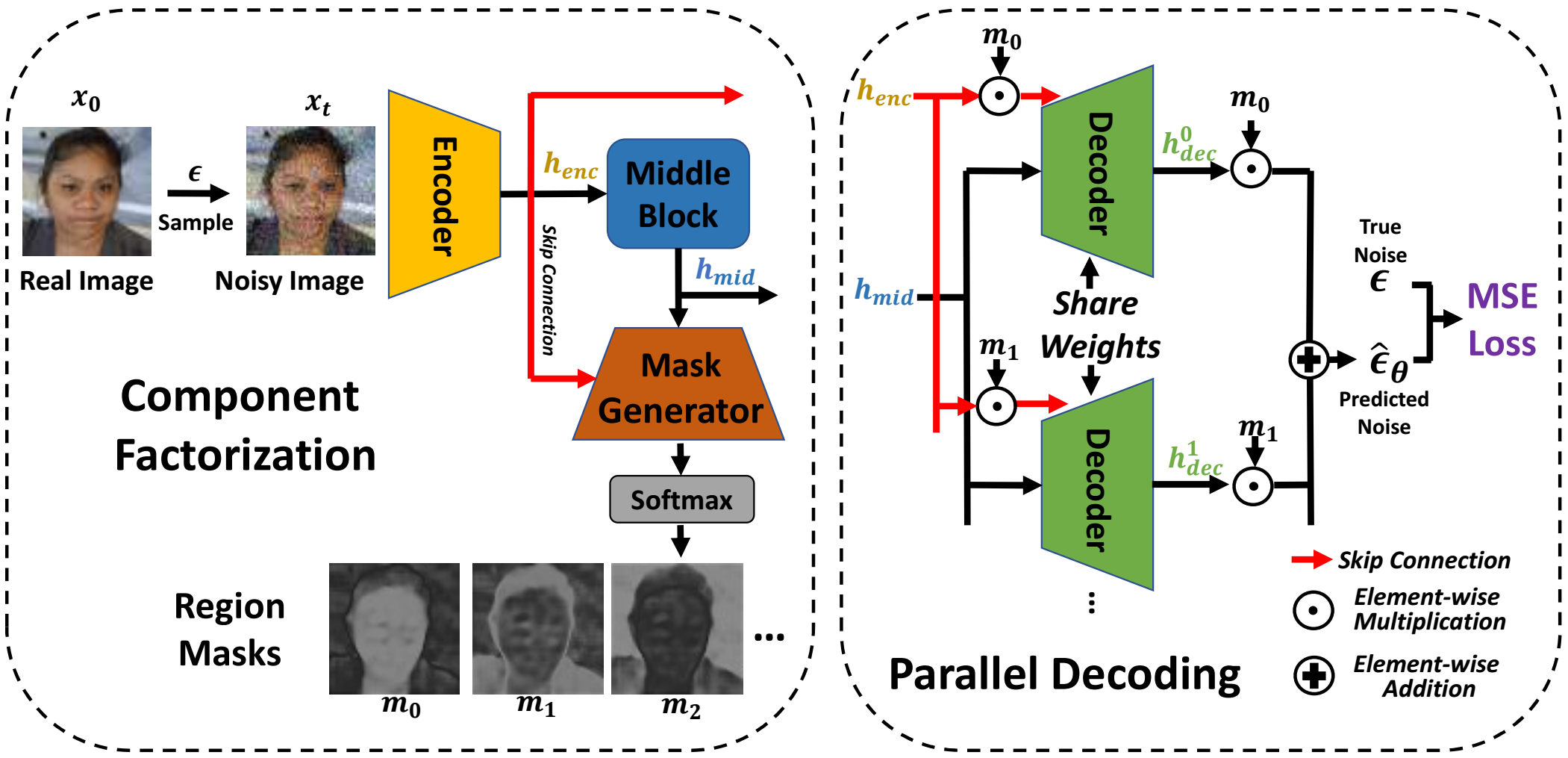}}
   \end{center}
   \vspace{-25pt}
   \caption{%
      \textbf{Factorized diffusion architecture.}
      Our framework restructures the architecture of the neural network within a DDPM~\cite{DBLP:conf/nips/HoJA20} so
      as to decompose the image denoising task into parallel subtasks.  All modules are end-to-end trainable and
      optimized according to the same denoising objective as DDPM.
      \emph{\textbf{Left:~Component factorization.}}
         An \emph{Encoder}, equivalent to the first half of a standard DDPM U-Net architecture, extracts features
         $\bh_{enc}$.  A common \emph{Middle Block} processes \emph{Encoder} output into shared latent features
         $\bh_{mid}$.  Note that \emph{Middle Block} and $\bh_{mid}$ exist in the standard denoising DDPM U-Net by
         default. We draw it as a standalone module for a better illustration of the detailed architectural design.
         A \emph{Mask Generator}, structured as the second half of a standard U-Net receives $\bh_{mid}$
         as input, alongside all encoder features $\bh_{enc}$ injected via skip connections to layers of corresponding
         resolution.  This later network produces a soft classification of every pixel into one of $K$ region masks,
         $\bm{m_0},\bm{m_1}, ..., \bm{m_K}$.
      \emph{\textbf{Right:~Parallel decoding.}}
         A \emph{Decoder}, also structured as the second half of a standard U-Net, runs separately for each region.
         Each instance of the \emph{Decoder} receives shared features $\bh_{mid}$ and a masked view of encoder
         features $\bh_{enc} \odot \bm{m_i}$ injected via skip connections to corresponding layers.  Decoder outputs
         are masked prior to combination.
	Though not pictured, we inject timestep embedding $\bm{t}$ into the \emph{Encoder}, \emph{Mask Generator}, and
      \emph{Decoder}.%
   }%
   \label{fig:framework}
\end{figure*}

Figure~\ref{fig:overview} provides an overview of our setting alongside example results, while
Figure~\ref{fig:framework} illustrates the details of our DDPM architecture which are fully presented in
Section~\ref{sec:method}.  This architecture constrains the computational resources available for denoising in a
manner that encourages learning of a factorized model of the data.  Specifically, each step of the DDPM has the
ability to utilize additional inference passes through multiple copies of a subnetwork if it is willing to decompose
the denoising task into parallel subproblems.  The specific decomposition strategy itself must be learned, but, by
design, is structured in a manner that reveals the solution to our target task of image segmentation.
We summarize our contributions as three-fold:
\begin{itemize}[leftmargin=.15in]
   \item{%
      \textbf{Unified learning of generation and segmentation.}~
      We train our new DDPM architecture once, obtaining a model directly applicable to two different
      tasks with zero modification or fine-tuning: image generation and image segmentation.
      Segmenting a novel input image is fast, comparable in speed to any system using a single forward
      pass of a U-Net~\cite{DBLP:conf/miccai/RonnebergerFB15} like architecture.%
   }%
   \item{%
      \textbf{Unsupervised segmentation for free.}~
      Our method automatically learns meaningful regions (\emph{e.g.,} foreground and background),
      guided only by the DDPM denoising objective; no extra regularization terms, no use of labels.%
   }%
   \item{%
      \textbf{Higher quality image synthesis.}~
      Our model generates higher-quality images than the baseline DDPM, as well as their corresponding
      segmentations simultaneously.  We achieve excellent quantitative and qualitative results under
      common evaluation protocols (Section~\ref{sec:experiments}).%
   }%
\end{itemize}

Beyond improvements to image generation and segmentation, our work can be viewed as the first case study of
a new paradigm for using generation as a learning objective, in combination with model architecture as a
constraint.  Rather than viewing a pre-trained generative model as a source from which to extract and
repurpose features for downstream tasks, design the model architecture in the first place so that, as a
byproduct of training from scratch to generate, it also learns to perform the desired task.

\section{Related Work}

\paragraph{Image Segmentation.}
Generic segmentation, which seeks to partition an image into meaningful regions without prior knowledge about object
categories present in the scene, is a longstanding challenge for computer vision.  Early methods rely on combinations
of hand-crafted features based on intensity, color, and texture cues~\cite{Canny:PAMI:1986,MFM:PAMI:2004}, clustering
algorithms~\cite{SM:PAMI:2000}, and a duality between closed contours and the regions they bound~\cite{AMFM:PAMI:2011}.
Deep learning modernized the feature representations used in these pipelines, yielding systems which, trained with
supervision from annotated regions~\cite{MFTM:ICCV:2001}, reach near human-level accuracy on predicting and localizing
region boundaries~\cite{deepedge,deepcontour,hed,kokkinos2015pushing}.

Semantic segmentation, which assigns a category label to each pixel location in image, has been similarly
revolutionized by deep learning.  Here, the development of specific architectures~\cite{DBLP:conf/cvpr/LongSD15,
DBLP:conf/miccai/RonnebergerFB15,DBLP:conf/cvpr/HariharanAGM15} enabled porting of approaches for image classification
to the task of semantic segmentation.

Recent research has refocused on the challenge of learning to segment without reliance on detailed annotation for
training.  Hwang~\etal~\cite{DBLP:conf/iccv/HwangYSCYZC19} combine two sequential clustering modules for
both pixel-level and segment-level to perform this task.  Ji~\etal~\cite{DBLP:conf/iccv/JiVH19} and
Ouali~\etal~\cite{DBLP:conf/eccv/OualiHT20} follow the concept of mutual information maximization to partition pixels
into two segments.  Savarese~\etal~\cite{DBLP:conf/cvpr/SavareseKMSM21} further propose a learning-free adversarial
method from the information theoretic perspective, with the goal of minimizing predictability among different pixel
subsets.

\noindent\textbf{Segmentation Learning in Generative Models.}
Generative model-based approaches produce a semantic mask by perturbing~\cite{DBLP:conf/nips/BielskiF19} or
redrawing~\cite{DBLP:conf/nips/ChenAD19} the generated foreground and background masks.  Despite good performance,
these methods apply only to two-class partitions and require extra loss terms based upon object priors in training
datasets.

Denoising diffusion probabilistic models (DDPMs)~\cite{DBLP:conf/nips/HoJA20} achieve state-of-the-art performance in
generating realistic images.  The incremental nature of their training process may offer advantages for scaling up
models in a stable manner.  A few recent works~\cite{DBLP:conf/iclr/BaranchukVRKB22,
DBLP:conf/cvpr/PreechakulCWS22,DBLP:conf/nips/ZhangZL22} explore representation learning capability in DDPMs.
DatasetDDPM~\cite{DBLP:conf/iclr/BaranchukVRKB22} is the first to explore the few-shot segmentation in pre-trained
diffusion models.  This method still requires human labels to train a linear classifier.  Moreover, with the default
U-Net architecture~\cite{DBLP:conf/miccai/RonnebergerFB15}, it loses the efficiency and flexibility of generating
image and masks in a single-stage manner.  Diff-AE~\cite{DBLP:conf/cvpr/PreechakulCWS22} and
PADE~\cite{DBLP:conf/nips/ZhangZL22} perform representation learning driven by a reconstruction objective in the
DDPM framework.  Unfortunately, their learned latent vector is not applicable to more challenging segmentation tasks.
Additionally, their learned representations require an extra pre-trained interpreter to perform downstream image
classification.

DiffuMask~\cite{DBLP:journals/corr/abs-2303-11681} takes a pre-trained Stable Diffusion
model~\cite{DBLP:conf/cvpr/RombachBLEO22}, which is trained with large-scale text-to-image datasets (and thus solves
a far less challenging problem), and conducts a post-hoc investigation on how to extract segmentation from its
attention maps.  Neither our system, nor the baseline DDPM to which we compare, makes use of such additional
information.  Furthermore, DiffuMask does not directly output segmentation; it is basically a dataset generator,
which produces generated images and pseudo labels, which are subsequently used to train a separate segmentation
model.  Our method, in contrast, is both completely unsupervised and provides an end-to-end solution by specifying
an architectural design in which training to generate reveals segmentations as a bonus.

MAGE~\cite{DBLP:journals/corr/abs-2211-09117} shares with us a similar motivation of framing generation and
representation learning in a unified framework.  However, our approach is distinct in terms of both
(1) task: we tackle a more complex unsupervised segmentation task (without fine-tuning) instead of image
classification (with downstream fine-tuning), and
(2) design: `masks' play a fundamentally different role in our system.
MAGE adopts an MAE~\cite{DBLP:conf/cvpr/HeCXLDG22}-like masking scheme on input data, in order to provide a proxy
reconstruction objective for self-supervised representation learning.  Our use of region masks serves a different
purpose, as they are integral components of the model being learned and facilitate factorization of the image
generation process into parallel synthesis of different segments.

\section{Factorized Diffusion Models}
\label{sec:method}

Figure~\ref{fig:framework} illustrates the overall architecture of our system, which partitions the denoising network
within a diffusion model into an unsupervised region mask generator and parallel per-region decoders.


\subsection{Unsupervised Region Factorization}

To simultaneously learn representations for both image generation and unsupervised segmentation, we first design the
region mask generator based on the first half (encoder) of a standard DDPM U-Net.  We obtain input $\bx_t$, a noised
version of $\bx_0$, via forward diffusion:
\begin{eqnarray} \label{eq:f_diff}
   q(\bx_t|\bx_0):= \mathcal{N}(\bx_t; \sqrt{\bar{\alpha}_t}\bx_0, (1-\bar{\alpha}_t)I), \nonumber \\
   \bx_{t} = \sqrt{\bar{\alpha}_t}\bx_0 + \sqrt{1-\bar{\alpha}_t}\bm{\epsilon}, \bm{\epsilon} \sim \mathcal{N}(0,1),
\end{eqnarray}
where $\alpha_t = 1 - \beta_t, \bar{\alpha}_t = \prod_{s=1}^t \alpha_t$.

In addition to the encoder half of the U-Net, we instantiate a middle block consisting of layers operating on lower
spatial resolution features.  Parameterizing these subnetworks as $\theta_{enc}$ and $\theta_{mid}$, we extract
latent representations:
\begin{eqnarray}
   \bh_{enc} = \theta_{enc}(\bx_t, t), \label{eq:enc} \\
   \bh_{mid} = \theta_{mid}({\bh_{enc}, t}) \label{eq:mid}
\end{eqnarray}
where $\bh_{enc}$ encapsulates features at all internal layers of $\theta_{enc}$, for subsequent use as inputs, via
skip connections, to corresponding layers of decoder-style networks (second half of a standard U-Net).

We instantiate a mask generator, $\theta_{mask}$, as one such decoder-style subnetwork.  A softmax layer produces an
output tensor with $K$ channels, representing $K$ different regions in image $\bx_0$:
\begin{eqnarray}\label{eq:mask_gen}
   \bm{m}_k = \theta_{mask}(\bh_{mid}, \bh_{enc}, t)
\end{eqnarray}
Following a U-Net architecture, $\bh_{enc}$ feeds into $\theta_{mask}$ through skip-connections.

\subsection{Parallel Decoding Through Weight Sharing}

We aim to extend a standard DDPM U-Net decoder $\theta_{dec}$ to consider region structure during generation.  One
simple design is to condition on $\bm{m} = \{\bm{m_0}, \bm{m_1}, ... \}$ by concatenating it with input
$\bh_{mid}$ and $\bh_{enc}$ along the channel dimension:
\begin{eqnarray}
   \hat{\bm{\epsilon}} = \theta_{dec}(\text{concat}[\bh_{mid},\bm{m}] , \text{concat}[\bh_{enc},\bm{m}], t),
\end{eqnarray}
where $\bh_{mid}$ and $\bh_{enc}$ are generated from Eq.~\ref{eq:enc} and Eq.~\ref{eq:mid}.
We downsample $\bm{m}$ accordingly to the same resolution as $\bh_{mid}$ and $\bh_{enc}$ at different stages.
However, such a design significantly modifies (\eg~channel sizes) the original U-Net decoder architecture.
Moreover, conditioning with the whole mask representation may also result in a trivial solution that simply
ignores region masks.

To address these issues, we separate the decoding scheme into multiple parallel branches of weight-shared U-Net
decoders, each masked by a single segment.  Noise prediction for $k$-th branch is:
\begin{eqnarray}
   \hat{\bm{\epsilon}}_k = \theta_{dec}(\bh_{mid},\bh_{enc} \odot \bm{m}_k, t)
\end{eqnarray}
and the output is a sum of region-masked predictions:
\begin{eqnarray}\label{eq:noise_pred}
   \hat{\bm{\epsilon}} = \sum_{k=0}^{K-1} \hat{\bm{\epsilon}}_k \odot \bm{m}_k
\end{eqnarray}

\begin{figure}[tp]
\begin{minipage}[t]{0.46\columnwidth}
\begin{algorithm}[H]
\footnotesize
\caption{\\Training Masked Diffusion}
\label{alg:training}
\begin{algorithmic}
  \STATE {\bfseries Input:} Data $\bm{x}_0$
  \STATE {\bfseries Output:} Trained model $\theta$
  \STATE {\bfseries Initialize:}\\{\quad}Model weights $\theta$,\\{\quad}Timesteps T
  \FOR{$\text{iter}=1$ {\bfseries to }Iter$_{total}$}
      \STATE Sample  $t \in [1,T]$
      \STATE Sample  $\bx_t$ using Eq.~\ref{eq:f_diff}
      \STATE Calculate $\hat{\bm{\epsilon}}$ using Eq.~\ref{eq:noise_pred}
      \STATE Backprop with Eq.~\ref{eq:loss}.
      \STATE Update $\theta$.
  \ENDFOR
  \STATE return $\theta$
\end{algorithmic}
\end{algorithm}
\end{minipage}
\hfill
\begin{minipage}[t]{0.52\columnwidth}
\begin{algorithm}[H]
\footnotesize
\caption{\\Image and Mask Generation}
\label{alg:eval}
\begin{algorithmic}
  \STATE {\bfseries Input:} Noise $\bx_T$, trained model $\theta$
  \STATE {\bfseries Output:}\\{\quad}Image $\hat{\bx_0}$ and segmentation $\hat{\bm{m}_0}$
  \STATE {\bfseries Initialize:} $\bx_T \sim \mathcal{N}(0,1)$
  \FOR{$\text{t}=T$ {\bfseries to } 1}
      \STATE Sample  $\bz$ using Eq.~\ref{eq:sample_z}
      \STATE Perform reversed diffusion using Eq.~\ref{eq:r_diff}
      \IF{$t = 1$}
      \STATE collect $\hat{\bm{m}_0}$ using Eq.~\ref{eq:mask_gen}
      \STATE return $\hat{\bx_0}$ and $\hat{\bm{m}_0}$. 
     \ENDIF
  \ENDFOR
\end{algorithmic}
\end{algorithm}
\end{minipage}
\end{figure}

\subsection{Optimization with Denoising Objective}

We train our model in an end-to-end manner, driven by the simple DDPM denoising objective.
Model weights $\theta=\{\theta_{enc},\theta_{mid}, \theta_{dec}, \theta_{mask}\}$ are optimized by
minimizing the noise prediction loss:
\begin{eqnarray} \label{eq:loss}
   L = \mathbb{E}||\bm{\epsilon} - \hat{\bm{\epsilon}}||_2^2
\end{eqnarray}
Unlike previous work, our method does not require a mask regularization loss term~\cite{DBLP:conf/cvpr/SavareseKMSM21,
DBLP:conf/nips/BielskiF19, DBLP:conf/nips/ChenAD19}, which pre-defines mask priors (\eg~object size).
Algorithm~\ref{alg:training} summarizes training.

\subsection{Segmentation via Reverse Diffusion}

Once trained, we can deploy our model to both segment novel input images as well as synthesize images from noise.

\noindent\textbf{Real Image Segmentation.}
Given clean input image $\bx_0$, we first sample a noisy version $\bx_t$ through forward diffusion in
Eq.~\ref{eq:f_diff}.  We then perform one-step denoising by passing $\bx_t$ to the model.  We collect the
predicted region masks as the segmentation for $\bx_0$ using Eq.~\ref{eq:mask_gen}.

\noindent\textbf{Image and Mask Generation.}
Using reverse diffusion, our model can generate realistic images and their corresponding segmentation masks,
starting from a pure noise input $\bx_T \sim \mathcal{N}(0,1)$.  Reverse diffusion predicts $\bx_{t-1}$ from $\bx_{t}$:
\begin{eqnarray}
   \bx_{t-1} = 1/\sqrt{\alpha_t}(\bx_t -\frac{1-\alpha_t}{\sqrt{1-\bar{\alpha_t}}}\theta(\bx_t,t))+\sigma_t\bz, \label{eq:r_diff}\\
   \bz \sim \mathcal{N}(0,1) \quad \text{if} \quad  t>1 \quad  \text{else} \quad  \bz=0. \label{eq:sample_z}
\end{eqnarray}
where  $\sigma_t$ is empirically set according to the DDPM noise scheduler.  We perform $T$ steps of reverse
diffusion to generate an image.  We also collect its corresponding mask using Eq.~\ref{eq:r_diff} when $t=1$.
Algorithm~\ref{alg:eval} summarizes this process.

\section{Experiments}
\label{sec:experiments}

We evaluate on: (1) real image segmentation, (2) image and region mask generation, using
Flower~\cite{DBLP:conf/icvgip/NilsbackZ08}, CUB~\cite{WahCUB_200_2011}, FFHQ~\cite{DBLP:conf/cvpr/KarrasLA19},
CelebAMask-HQ~\cite{CelebAMask-HQ}, and ImageNet~\cite{IMAGENET}.

\noindent \textbf{Evaluation Metrics.}
For unsupervised segmentation on Flower and CUB, we follow the data splitting in
IEM~\cite{DBLP:conf/cvpr/SavareseKMSM21} and evaluate predicted mask quality under three commonly used metrics,
denoted as Acc., IOU and DICE score~\cite{DBLP:conf/cvpr/SavareseKMSM21,DBLP:conf/nips/ChenAD19}.
Acc.~is the (per-pixel) mean accuracy of the foreground prediction.
IOU is the predicted foreground region's intersection over union (IoU) with the ground-truth foreground region.
DICE score is defined as $2\frac{\hat{F}\cap F}{|\hat{F}|}$~\cite{dice1945measures}.
On ImageNet, we evaluate our method on Pixel-ImageNet~\cite{zhang2020interactive}, which provides human-labeled
segmentation masks for 0.485M images covering 946 object classes.  We report Acc., IOU and DICE score on a randomly
sampled subset, each class containing at most 20 images.  For face datasets, we train our model on FFHQ and only
report per-pixel accuracy on the CelebAMask test set, using provided ground-truth.

For image and mask generation, we use Fr{\'{e}}chet Inception Distance (FID)~\cite{DBLP:conf/nips/HeuselRUNH17}
for generation quality assessment.  Since we can not obtain the ground-truth for generated masks, we apply
a supervised U-Net segmentation model, pre-trained on respective datasets, to the generated images and measure the
consistency between masks in terms of per-pixel accuracy.  In addition to quantitative comparisons, we show
extensive qualitative results.

\noindent\textbf{Implementation Details.}
We train Flower, CUB and Face models at both $64\times 64$ and $128\times 128$ resolution.
We also train class-conditioned ImageNet models with $64\times 64$ resolution. 
For all experiments, we use the U-Net~\cite{DBLP:conf/miccai/RonnebergerFB15} encoder-middle-decoder architecture
similar to~\cite{DBLP:conf/nips/HoJA20}.  We use the decoder architecture as our mask generator and set the number
of factorized masks $K$ as 3.
For $64 \times 64$ the architecture is as follows:
The downsampling stack performs four steps of downsampling, each with 3 residual blocks.
The upsampling stack is setup as a mirror image of the downsampling stack.
From highest to lowest resolution, U-Net stages use $[C,2C,3C,4C]$ channels, respectively.
For $128 \times 128$ architecture, the down/up sampling block is 5-step with $[C,C,2C,3C,4C]$ channels, each with two
residual blocks, respectively.  We set $C = 128$ for all models.

We use Adam to train all the models with a learning rate of $10^{-4}$ and an exponential moving average (EMA) over
model parameters with rate $0.9999$.  For all datasets except ImageNet, we train $64\times 64$ and $128\times 128$
models on 8 and 32 Nvidia V100 32GB GPUS, respectively.  For Flower, CUB and FFHQ, we train the models for 50K, 50K,
500K iterations with batch size of 128, respectively.  For ImageNet, we train 500K iterations on 32 Nvidia V100 GPUS
with batch size 512.  We adopt the linear noise scheduler as in Ho~\etal~\cite{DBLP:conf/nips/HoJA20} with $T=1000$ timesteps.

\subsection{Image Segmentation}
\begin{figure}
  \begin{center}
  \begin{minipage}[tbh]{\linewidth}
    \centering
   \subfloat[\textbf{\textsf{\scriptsize{Real Images}}}] {
      \includegraphics[width = 0.46\columnwidth]{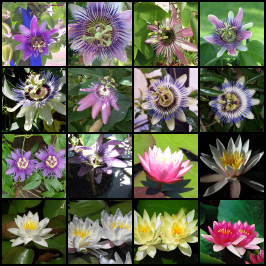}
   }
   \subfloat[\textbf{\textsf{\scriptsize{Segmentation}}}] {
      \includegraphics[width = 0.46\columnwidth]{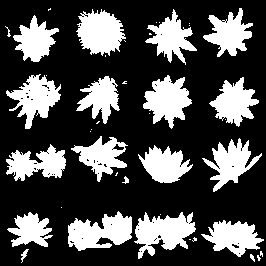}
   }
   \vspace{-8pt}
    \caption
      {%
        {Segmentation on Flower.}
        \label{fig:seg_flower}%
      }%
  \end{minipage}%
  \vspace{10pt}
  \begin{minipage}[tbh]{\linewidth}
    \centering
    \begin{tabular}{lccccccccc}
    \toprule
    &\multicolumn{1}{c}{{Methods}}
    &\multicolumn{1}{c}{Acc.}
    &\multicolumn{1}{c}{IOU}
    &\multicolumn{1}{c}{DICE} \\
    \midrule
    &GrabCut~\cite{DBLP:journals/tog/RotherKB04} &82.0 &69.2 &79.1\\
    &ReDO~\cite{DBLP:conf/nips/ChenAD19} &87.9 &76.4 &-\\
    &IEM~\cite{DBLP:conf/cvpr/SavareseKMSM21} &88.3 &76.8 &84.6\\
    &IEM+SegNet~\cite{DBLP:conf/cvpr/SavareseKMSM21} &89.6 &78.9 &86.0\\
    \midrule
    &Ours &\textbf{90.1} &\textbf{79.7} &\textbf{87.2} \\
    \bottomrule
    \end{tabular}
    \vspace{-6pt}
    \captionof{table}{{Comparisons on Flower.}}
    \label{tab:seg_flower}
      \end{minipage}
   \end{center}
   \vspace{-20pt}
\end{figure}

\begin{figure}
  \begin{center}
  \begin{minipage}[tbh]{\linewidth}
    \centering
   \subfloat[\textbf{\textsf{\scriptsize{Real Images}}}] {
      \includegraphics[width = 0.46\columnwidth]{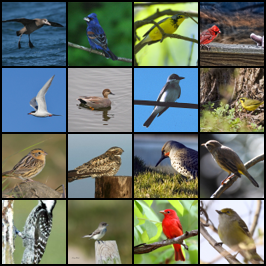}
   }
   \subfloat[\textbf{\textsf{\scriptsize{Segmentation}}}] {
      \includegraphics[width = 0.46\columnwidth]{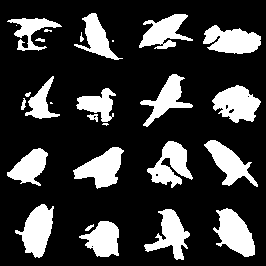}
   }
   \vspace{-8pt}
    \caption
      {%
        {Segmentation on CUB.}
        \label{fig:seg_cub}%
      }%
  \end{minipage}%
  \vspace{10pt}
  \begin{minipage}[tbh]{\linewidth}
    \centering
    \begin{tabular}{lccccccccc}
    \toprule
    &\multicolumn{1}{c}{{Methods}}
    &\multicolumn{1}{c}{Acc.}
    &\multicolumn{1}{c}{IOU}
    &\multicolumn{1}{c}{DICE} \\
    \midrule
    &GrabCut~\cite{DBLP:journals/tog/RotherKB04} &72.3 &36.0 &48.7 \\
    &PerturbGAN~\cite{DBLP:conf/nips/BielskiF19}&- &38.0 &- \\
    &ReDO~\cite{DBLP:conf/nips/ChenAD19} &84.5 &42.6 &-\\
    &IEM~\cite{DBLP:conf/cvpr/SavareseKMSM21} &88.6 &52.2 &66.0\\
    &IEM+SegNet~\cite{DBLP:conf/cvpr/SavareseKMSM21} &89.3 &55.1 &68.7\\
    \midrule
    &Ours &\textbf{89.6} &\textbf{56.1} &\textbf{69.4} \\
    \bottomrule
    \end{tabular}
    \vspace{-6pt}
    \captionof{table}{{Comparisons on CUB.}} 
    \label{tab:seg_cub}
      \end{minipage}
   \end{center}
   \vspace{-15pt}
\end{figure}

To evaluate our method on real image segmentation, we set $t$ as 30 for forward diffusion process.
For Flower and CUB, Figures~\ref{fig:seg_flower} and~\ref{fig:seg_cub} show test images and predicted segmentations.
Tables~\ref{tab:seg_flower} and~\ref{tab:seg_cub} provide quantitative comparison with representative unsupervised
image segmentation methods:
GrabCut~\cite{DBLP:journals/tog/RotherKB04},
ReDO~\cite{DBLP:conf/nips/ChenAD19}
and IEM~\cite{DBLP:conf/cvpr/SavareseKMSM21}.
As shown in Table~\ref{tab:seg_flower} and Table~\ref{tab:seg_cub}, our method outperforms all competitors.

\begin{figure}
  \begin{center}
  \begin{minipage}[tbh]{\linewidth}
    \centering
   \subfloat[\textbf{\textsf{\scriptsize{Real Images}}}] {
      \includegraphics[width = 0.46\columnwidth]{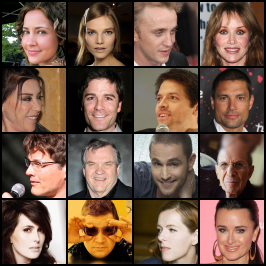}
   }
   \subfloat[\textbf{\textsf{\scriptsize{Segmentation}}}] {
      \includegraphics[width = 0.46\columnwidth]{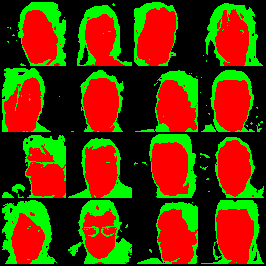}
   }
   \vspace{-8pt}
    \caption
      {%
        {Segmentation on CelebA.}
        \label{fig:seg_celebA}%
      }%
  \end{minipage}%
  \vspace{10pt}
  \begin{minipage}[tbh]{\linewidth}
    \centering
    \begin{tabular}{lccccccccc}
    \toprule
    &\multicolumn{1}{c}{{Methods}}
    &\multicolumn{1}{c}{Acc.}
    &\multicolumn{1}{c}{mIOU} \\
    \midrule
    &Supervised UNet &95.7 &90.2 \\
    \midrule
    &DatasetDDPM-unsup.~\cite{DBLP:conf/iclr/BaranchukVRKB22}&78.5 &69.3  \\
    \midrule
    &Ours &87.9 &80.3\\
    \bottomrule
    \end{tabular}
    \vspace{-6pt}
    \captionof{table}{{Segmentation comparisons on CelebA.}}
    \label{tab:seg_celebA}
      \end{minipage}
   \end{center}
   \vspace{-20pt}
\end{figure}

We also visualize the predicted face parsing results on FFHQ and CelebAMask datasets in Figure~\ref{fig:overview}(c)(d)
and Figure~\ref{fig:seg_celebA}.  Our model learns to accurately predict three segments corresponding to semantic
components: skin, hair, and background.  This particular semantic partitioning emerges from our unsupervised learning
objective, without any additional prior.  With ground-truth provided on CelebAMask-HQ, we also compare the pixel
accuracy and mean of IOU with a supervised U-Net and DatasetDDPM~\cite{DBLP:conf/iclr/BaranchukVRKB22}.  For the
former, we train a supervised segmentation model with 3-class cross-entropy loss.  For the unsupervised setting, we
perform K-means (K=3) on the pre-trained DDPM, denoted as DatasetDDPM-unsup.  Table~\ref{tab:seg_celebA} shows that we
outperform DatasetDDPM by a large margin and achieve a relatively small performance gap with a supervised U-Net.

\begin{figure}
  \begin{minipage}[tbh]{\linewidth}
    \centering
   \subfloat[\textbf{\textsf{\scriptsize{Real Images}}}] {
      \includegraphics[width = 0.46\columnwidth]{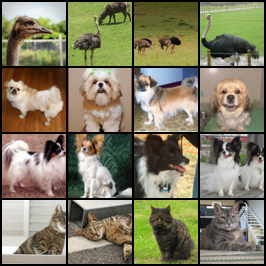}
   }
   \subfloat[\textbf{\textsf{\scriptsize{Segmentation}}}] {
      \includegraphics[width = 0.46\columnwidth]{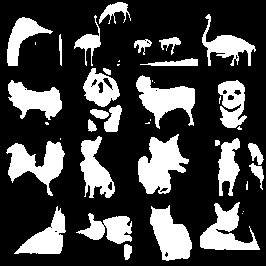}
   }
   \vspace{-8pt}
    \caption
      {%
        {Segmentation on ImageNet.}
        \label{fig:seg_imgnet}%
      }%
  \end{minipage}%
  \vspace{10pt}
  \begin{minipage}[tbh]{\linewidth}
    \centering
    \begin{tabular}{lccccccccc}
    \toprule
    &\multicolumn{1}{c}{{Methods}}
    &\multicolumn{1}{c}{Acc.}
    &\multicolumn{1}{c}{mIOU} \\
    \midrule
    &Supervised UNet &85.7 &74.1 \\
    \midrule
    &DatasetDDPM-unsup.~\cite{DBLP:conf/iclr/BaranchukVRKB22} &74.1 &60.4 \\
    \midrule
    &Ours &80.7 &67.7 \\
    \bottomrule
    \end{tabular}
    \vspace{-6pt}
    \captionof{table}{{Segmentation comparisons on ImageNet.}}
    \label{tab:seg_imgnet}
      \end{minipage}
      \vspace{-10pt}
\end{figure}

Figure~\ref{fig:seg_imgnet} shows the accurate segmentation results for ImageNet classes: ostrich, pekinese,
papillon, and tabby.  We compare with supervised U-Net and DatasetDDPM-unsup in
Table~\ref{tab:seg_imgnet}. We show more visualizations in Appendix Section~\ref{appendix:gen}.

\subsection{Image and Mask Generation}
\begin{figure}
  \begin{minipage}[tbh]{\linewidth}
    \centering
   \subfloat[\textbf{\textsf{\scriptsize{Generated Images}}}] {
      \includegraphics[width = 0.46\columnwidth]{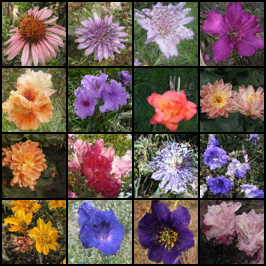}
   }
   \subfloat[\textbf{\textsf{\scriptsize{Generated Masks}}}] {
      \includegraphics[width = 0.46\columnwidth]{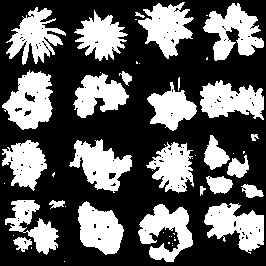}
   }
  \vspace{-8pt}
    \caption
      {%
        Generation on Flower.
        \label{fig:sample_flower}%
      }%
  \end{minipage}%
  \vspace{10pt}
  \begin{minipage}[tbh]{\linewidth}
    \centering
   \subfloat[Generated Images.] {
      \includegraphics[width = 0.46\columnwidth]{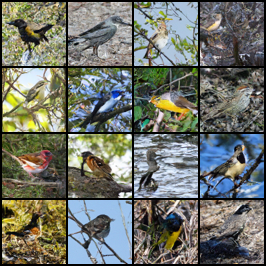}
   }
   \subfloat[Generated Masks.] {
      \includegraphics[width = 0.46\columnwidth]{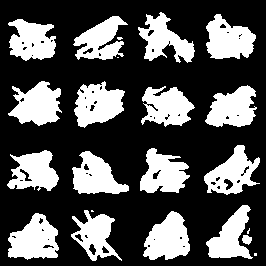}
   }
  \vspace{-8pt}
    \caption
      {%
        Generation on CUB.
        \label{fig:sample_cub}%
      }%
  \end{minipage}
\end{figure}

We evaluate our method on image and mask generation.
As shown in Figure~\ref{fig:sample_flower},~\ref{fig:sample_cub},~\ref{fig:overview}(c)(d) and~\ref{fig:sample_imgnet},
our method is able to generate realistic images.  In the upper row of Table~\ref{tab:genall:result}, we see a consistent
quality improvement over the original DDPM.  This suggests our method as a better architecture than standard U-Net
through separating computational power to each individual image segment, which may benefit the denoising task during
training.  More importantly, our method can produce accurate corresponding masks, closely aligned with the
semantic partitions in the generated image.

We evaluate the segmentation quality. Since there is no ground-truth mask provided for generated images, we apply
the U-Net segmentation models (pre-trained on respective labeled training sets) to the generated images to produce
reference masks.  We measure the consistency between the reference and the predicted parsing results in terms of
pixel-wise accuracy.  We compare our method with a pre-trained DDPM baseline, in which we first perform image
generation, then pass them to DatasetDDPM-unsup~to get masks.  As shown in Table~\ref{tab:genall:result} (bottom), our
method consistently achieves better segmentation on generated images than the DDPM baseline.  Note that, different
from the two-stage baseline, our method performs the computation in a single stage, generating image and mask
simultaneously.  Appendix Section~\ref{appendix:gen} shows more visualizations.

\begin{table*}[tb]
\begin{center}
\caption{Image and mask generation comparison on all datasets. (upper: FID($\downarrow$) bottom: Acc. ($\uparrow$))}
\label{tab:genall:result}
\vspace{-10pt}
\setlength{\tabcolsep}{10pt}
\begin{tabular}
{lccccccccc}
\toprule
&\multicolumn{1}{c}{{Models}}
&\multicolumn{1}{c}{Flower-64}
&\multicolumn{1}{c}{Flower-128}
&\multicolumn{1}{c}{CUB-64}
&\multicolumn{1}{c}{CUB-128}
&\multicolumn{1}{c}{FFHQ-64}
&\multicolumn{1}{c}{FFHQ-128}
&\multicolumn{1}{c}{ImageNet-64}\\
\midrule
&DDPM &15.81 &14.62 &14.45 &14.01 &13.72 &13.35 &7.02\\
&Ours &\textbf{13.33} &\textbf{11.50} &\textbf{10.91} &\textbf{10.28} &\textbf{12.02} &\textbf{10.79} &\textbf{6.54}\\
\midrule
&DDPM &80.5 &82.9 &84.2 &83.7 &84.2 &84.2 &71.2 \\
&Ours &\textbf{92.3} &\textbf{92.7} &\textbf{91.4} &\textbf{91.2} &\textbf{90.3} &\textbf{90.7} &\textbf{84.1} \\
\bottomrule
\end{tabular}
\end{center}
\vspace{-2em}
\end{table*}

\subsection{Ablation Study and Analysis}

\begin{figure}
  \begin{minipage}[tbh]{\linewidth}
    \centering
    \includegraphics[width = \columnwidth]{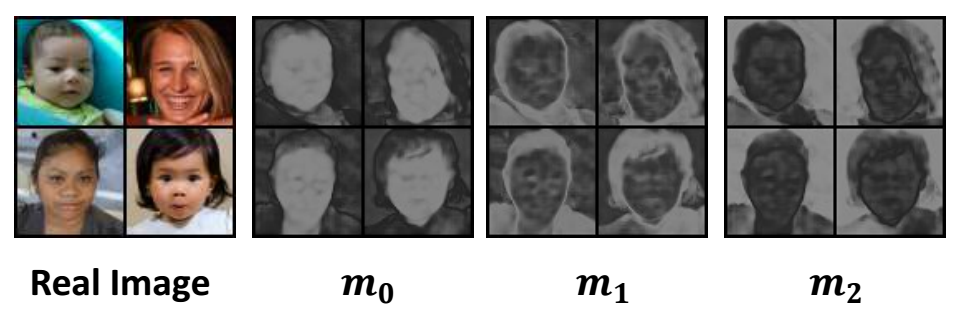}
    \vspace{-20pt}
    \caption
      {%
        Mask factorization (3 parts) on FFHQ.
        \label{fig:sample_factorization}%
      }%
  \end{minipage}%
  \vspace{10pt}
  \begin{minipage}[tbh]{\linewidth}
    \centering
      \includegraphics[width = \columnwidth]{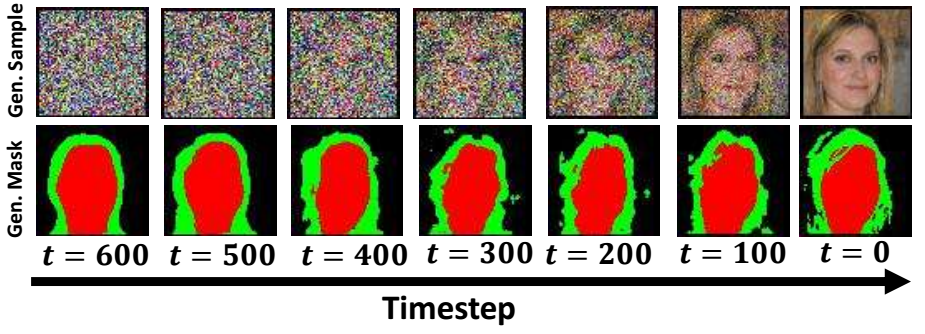}
   \vspace{-20pt}
    \caption
      {%
        Generation refinement along diffusion.
        \label{fig:sample_refinement}%
      }%
  \end{minipage}
  \vspace{-5pt}
\end{figure}

\begin{figure}
  \begin{minipage}[tbh]{\linewidth}
    \centering
   \subfloat[\textbf{\textsf{\scriptsize{Generated Images}}}] {
      \includegraphics[width = 0.46\columnwidth]{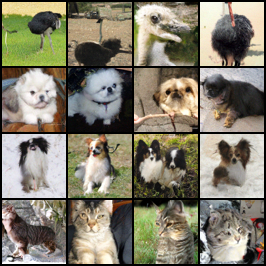}
   }
   \subfloat[\textbf{\textsf{\scriptsize{Generated Masks}}}] {
      \includegraphics[width = 0.46\columnwidth]{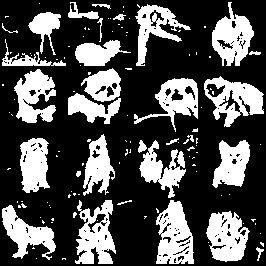}
   }
   \vspace{-8pt}
    \caption
      {%
        Conditional generation on ImageNet.
        \label{fig:sample_imgnet}%
      }%
  \end{minipage}%
  \vspace{10pt}
  \begin{minipage}[tbh]{\linewidth}
    \centering
    \begin{tabular}{lccccccccc}
    \toprule
    &\multicolumn{1}{c}{{Methods}}
    &\multicolumn{1}{c}{IOU.($\uparrow$)}
    &\multicolumn{1}{c}{FID ($\downarrow$)} \\
    \midrule
    &Concat &20.7 &14.21 \\
    &Masking $\bh_{mid}$ &20.2 &14.33 \\
    &w/o weight sharing &50.5 &17.21 \\
    \midrule
    &Ours &\textbf{56.1} &\textbf{10.28} \\
    \bottomrule
    \end{tabular}
    \vspace{-6pt}
    \captionof{table}{Ablations of decoding scheme on CUB.}
    \label{tab:abla_decoder}
      \end{minipage}
      \vspace{-10pt}
\end{figure}

\begin{figure}[t]
\begin{center}
\centerline{\includegraphics[width=\linewidth]{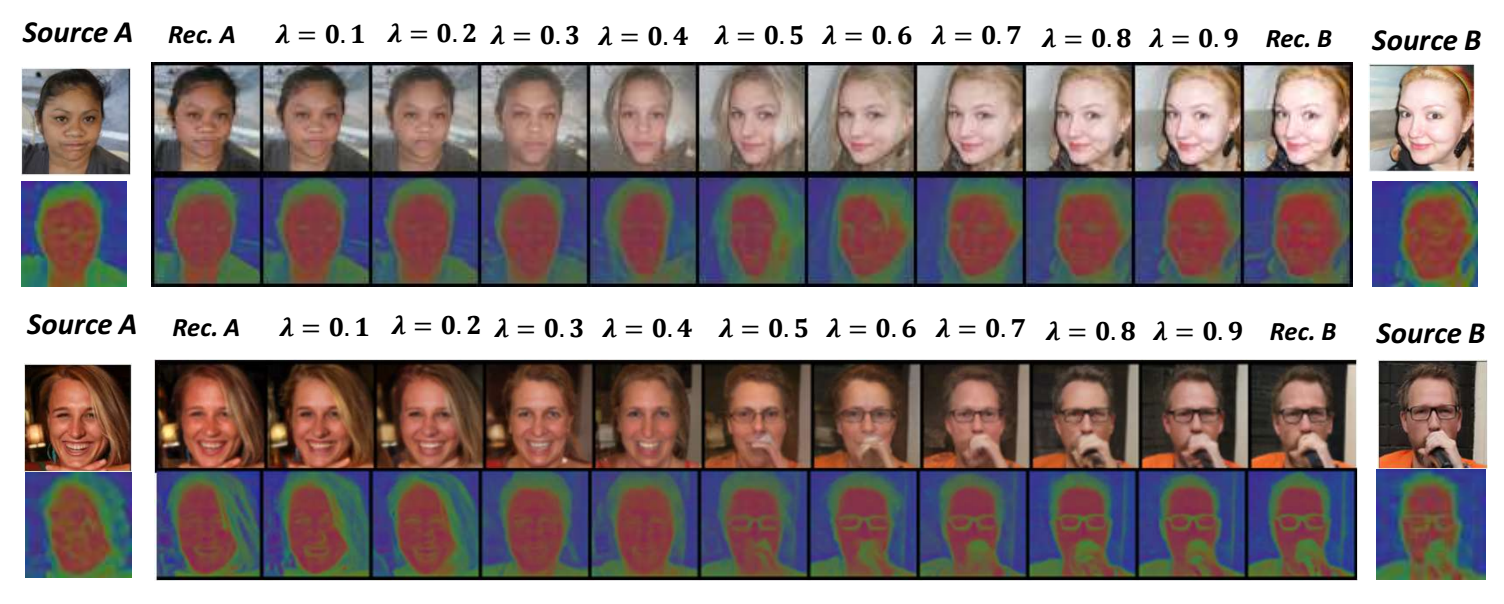}}
\vspace{-10pt}
\caption{Interpolations of FFHQ with 250 timesteps of diffusion.
}
\label{fig:interpolate}
\end{center}
\vspace{-20pt}
\end{figure}
\noindent\textbf{Multi-branch Decoders with Weight Sharing.}~
Separating computation in multi-branch decoders with weight sharing is an essential design in our method.
We show the effectiveness of this design by varying how to apply factorized masks in our decoding scheme:
(1) concat: we use single branch to take concatenation of $\bh$ and $\bm{m}$.
(2) masking $\bh_{mid}$: we use $\bm{m}$ to mask $\bh_{mid}$ instead of $\bh_{enc}$.
(3) w/o weight sharing: we train decoders separately in our design.
Table~\ref{tab:abla_decoder} shows separate design consistently yields better visual features than other designs for
CUB.
This suggests that our design benefits from fully utilizing mask information in the end-to-end denoising task and
avoids a trivial solution where masks are simply ignored.

\noindent\textbf{Investigation on Mask Factorization.}~
Our architecture is able to generate factorized representations, each representing a particular segment of the input
image.  We show this by visualizing the individual channels from softmax layer output in our mask generator.
As shown in Figure~\ref{fig:sample_factorization}, skin, hair, and background are separated in different channels.

\noindent\textbf{Mask Refinement along Diffusion Process.}~
In the DDPM Markov process, the model implicitly formulates a mapping between noise and data distributions.
We validate that this occurs for both images and latent region masks by visualizing image and mask generation
along the sequential reversed diffusion process in Figure~\ref{fig:sample_refinement}.  We observe gradual
refinement as denoising steps approach $t=0$.

\noindent\textbf{Face Interpolation.}~
We also investigate the face interpolation task on FFHQ.  Similar to standard DDPM~\cite{DBLP:conf/nips/HoJA20}, we
perform the interpolation in the denoising latent space with 250 timesteps of diffusion.  Figure~\ref{fig:interpolate}
shows good reconstruction in both pixels and region masks, yielding smoothly varying interpolations across face
attributes such as pose, skin, hair, expression, and background.

\begin{figure}[t]
\begin{subfigure}[b]{\linewidth}
\centering
\includegraphics[width=\textwidth]{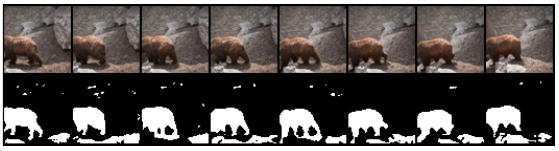}
\caption{\textbf{\textsf{\scriptsize{Frames of `Bear'}}}}
\label{seg:daivs_bear}
\end{subfigure}
\begin{subfigure}[b]{\linewidth}
\centering
\includegraphics[width=\textwidth]{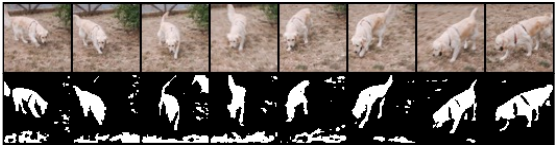}
\caption{\textbf{\textsf{\scriptsize{Frames of `Dog'}}}}
\label{seg:daivs_dog}
\end{subfigure}
\vspace{-20pt}
\caption
  {%
    {Segmentation on DAVIS-2017.}
    \label{fig:seg_davis}%
  }%
\end{figure}

\noindent\textbf{Zero-shot Object Segmentation.}~
We evaluate zero-shot object segmentation on both PASCAL VOC 2012~\cite{pascal-voc-2012} and DAVIS-2017
videos~\cite{Pont-Tuset_arXiv_2017}.  Baseline DDPM generation is not solved for these datasets when training
from scratch without external large-scale datasets (\emph{e.g.,} LAION~\cite{DBLP:conf/nips/SchuhmannBVGWCC22} used
in Stable Diffusion~\cite{DBLP:conf/cvpr/RombachBLEO22}).  We directly adopt zero-shot transfer from our pre-trained
ImageNet model by applying the conditional label mapping.  We detail the mapping rule in Appendix Section~\ref{appendix:mapping}.
Figure~\ref{fig:seg_voc} shows the accurate segmentation results for images of classes: aeroplane, monitor, person,
and sofa from VOC.  Since our method does not require any pixel labels, we evaluate the performance of each object
class individually.  Our method achieves a favorable high accuracy of $\textbf{0.78}$ and mIOU of $\textbf{0.54}$ when
averaging over all 20 classes.  We also report the results for each individual class in Appendix Section~\ref{appendix:voc}.  We also show
video segmentation on DAVIS-2017 in Figure~\ref{fig:seg_davis} and Appendix Section~\ref{appendix:davis}, without any labeled video pre-training.

\begin{figure}[t]
    \centering
   \subfloat[\textbf{\textsf{\scriptsize{Real Images}}}] {
      \includegraphics[width = 0.46\columnwidth]{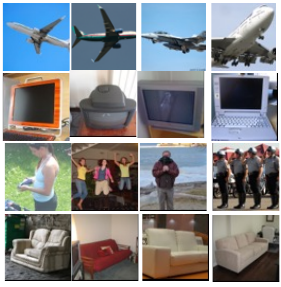}
   }
   \subfloat[\textbf{\textsf{\scriptsize{Segmentation}}}] {
      \includegraphics[width = 0.46\columnwidth]{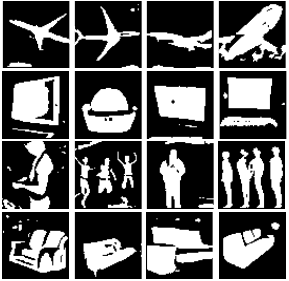}
   }
   \vspace{-8pt}
    \caption
      {%
        {Segmentation on VOC-2012.}
        \label{fig:seg_voc}%
      }%
\end{figure}

\section{Conclusion}

We propose a factorized architecture for diffusion models that is able to perform unsupervised image segmentation and
generation simultaneously, while being trained once, from scratch, for image generation via denoising alone.  Using
model architecture as a constraint, via carefully designed component factorization and parallel decoding schemes, our
method effectively and efficiently bridges these two challenging tasks in a unified framework, without the need of
fine-tuning or alternating the original DDPM training objective.  Our work is the first example of engineering an
architectural bottleneck so that learning a desired end task becomes a necessary byproduct of training to generate.

{
    \small
    \bibliographystyle{ieeenat_fullname}
    \bibliography{ref}
}

\newpage
\section{Appendix}

\subsection{Additional Segmentation Results}\label{appendix:seg}
\begin{figure*}[b]
    \centering
   \subfloat[\textbf{\textsf{\scriptsize{Real Images}}}] {
      \includegraphics[width = \columnwidth]{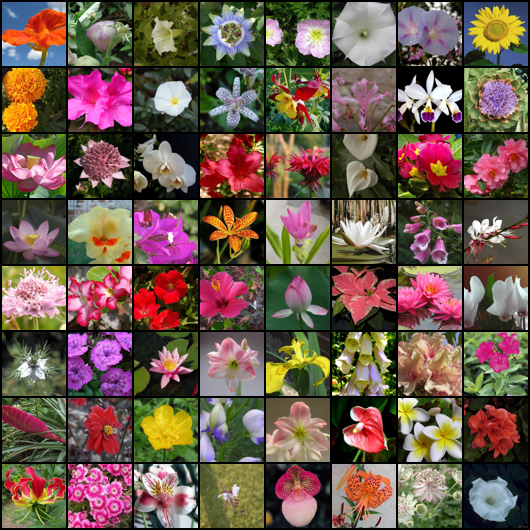}
   }
   \subfloat[\textbf{\textsf{\scriptsize{Segmentation}}}] {
      \includegraphics[width = \columnwidth]{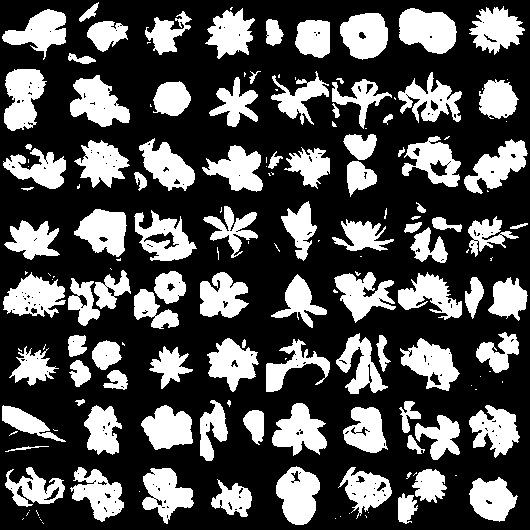}
   }
    \caption
      {%
        {Segmentation on Flower.}
        \label{fig:append:seg_flower}%
      }%
  \end{figure*}

\begin{figure*}[htbp]
    \centering
   \subfloat[\textbf{\textsf{\scriptsize{Real Images}}}] {
      \includegraphics[width = \columnwidth]{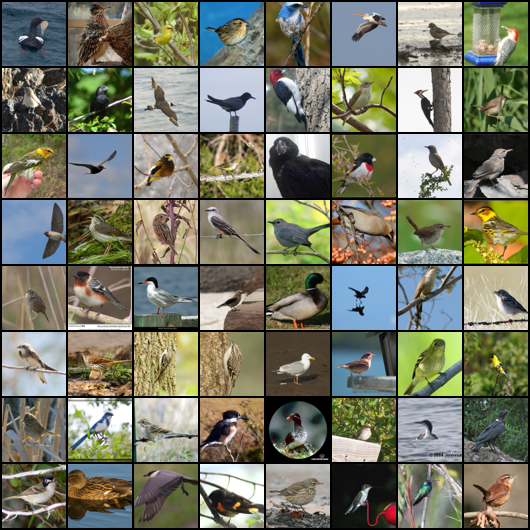}
   }
   \subfloat[\textbf{\textsf{\scriptsize{Segmentation}}}] {
      \includegraphics[width = \columnwidth]{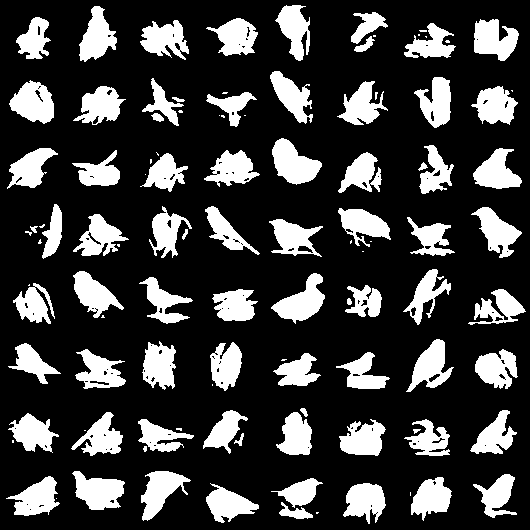}
   }
    \caption
      {%
        {Segmentation on CUB.}
        \label{fig:append:seg_cub}%
      }%
  \end{figure*}

\begin{figure*}[htbp]
    \centering
   \subfloat[\textbf{\textsf{\scriptsize{Real Images}}}] {
      \includegraphics[width = \columnwidth]{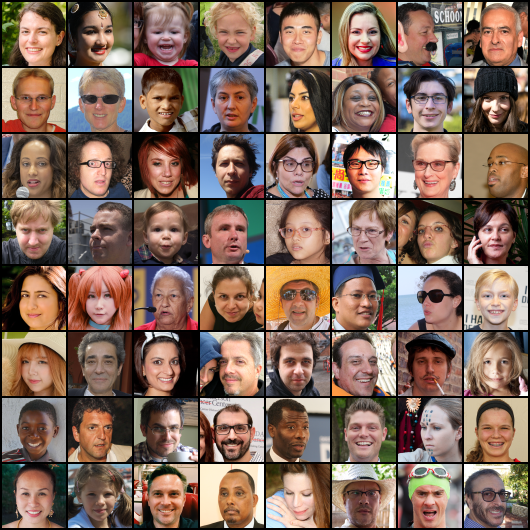}
   }
   \subfloat[\textbf{\textsf{\scriptsize{Segmentation}}}] {
      \includegraphics[width = \columnwidth]{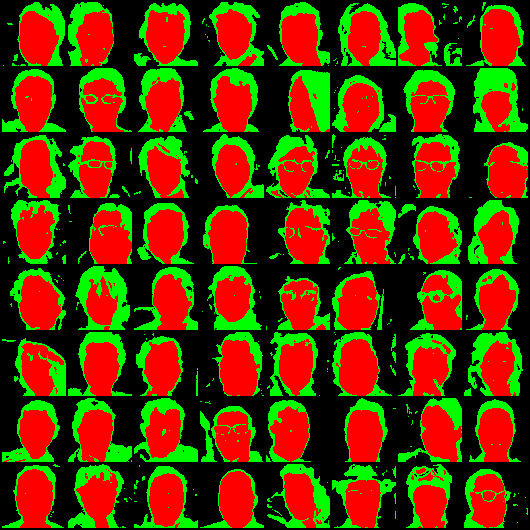}
   }
    \caption
      {%
        {Segmentation on FFHQ.}
        \label{fig:append:seg_ffhq}%
      }%
  \end{figure*}

\begin{figure*}[htbp]
    \centering
   \subfloat[\textbf{\textsf{\scriptsize{Real Images}}}] {
      \includegraphics[width = \columnwidth]{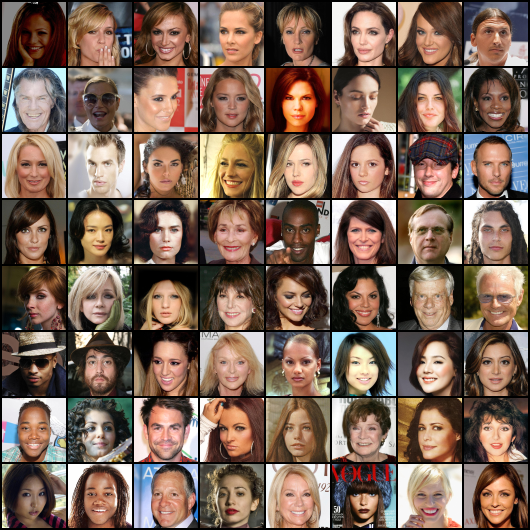}
   }
   \subfloat[\textbf{\textsf{\scriptsize{Segmentation}}}] {
      \includegraphics[width = \columnwidth]{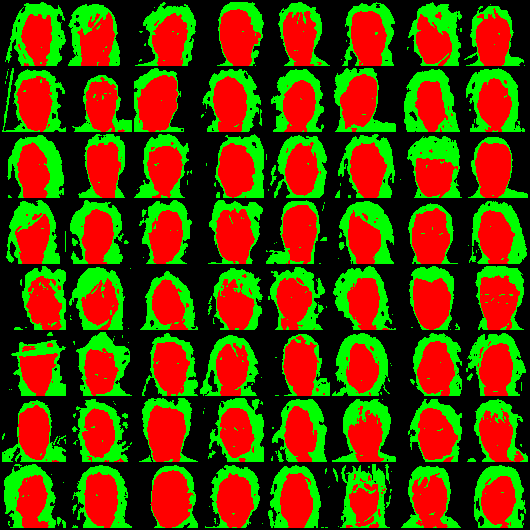}
   }
    \caption
      {%
        {Segmentation on CelebA.}
        \label{fig:append:seg_celebA}%
      }%
  \end{figure*}

\begin{figure*}[htbp]
    \centering
   \subfloat[\textbf{\textsf{\scriptsize{Real Images}}}] {
      \includegraphics[width = \columnwidth]{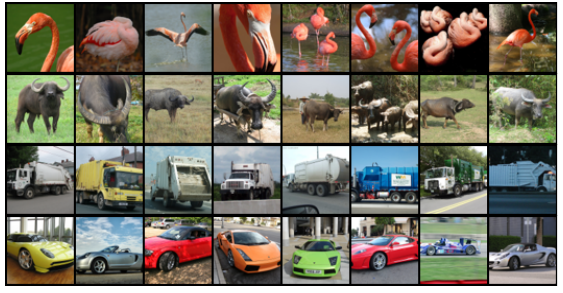}
   }
   \subfloat[\textbf{\textsf{\scriptsize{Segmentation}}}] {
      \includegraphics[width = \columnwidth]{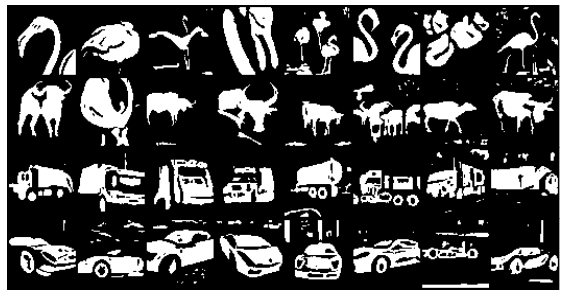}
   }
    \caption
      {%
        {Segmentation on ImageNet.}
        \label{fig:append:seg_imgnet}%
      }%
  \end{figure*}

We show more segmentation results for Flower, CUB, FFHQ, CelebA and ImageNet. As shown in Figures~\ref{fig:append:seg_flower},~\ref{fig:append:seg_cub},~\ref{fig:append:seg_ffhq},~\ref{fig:append:seg_celebA}, and~\ref{fig:append:seg_imgnet}, our method consistently predicts accurate segmentations for real image inputs.

\subsection{Additional Generation Results}\label{appendix:gen}
\begin{figure*}[htbp]
    \centering
   \subfloat[\textbf{\textsf{\scriptsize{Generated Images}}}] {
      \includegraphics[width = \columnwidth]{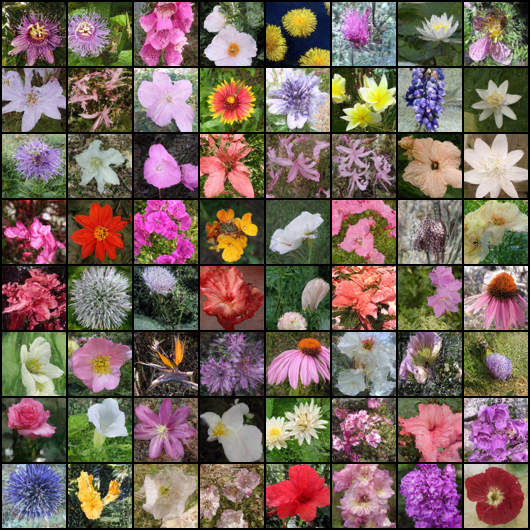}
   }
   \subfloat[\textbf{\textsf{\scriptsize{Generated Masks}}}] {
      \includegraphics[width = \columnwidth]{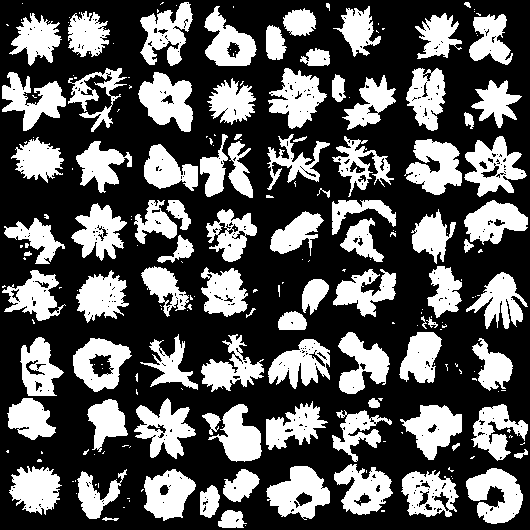}
   }
    \caption
      {%
        Generation on Flower.
        \label{fig:append:gen_flower}%
      }%
\end{figure*}

\begin{figure*}[htbp]
    \centering
   \subfloat[\textbf{\textsf{\scriptsize{Generated Images}}}] {
      \includegraphics[width = \columnwidth]{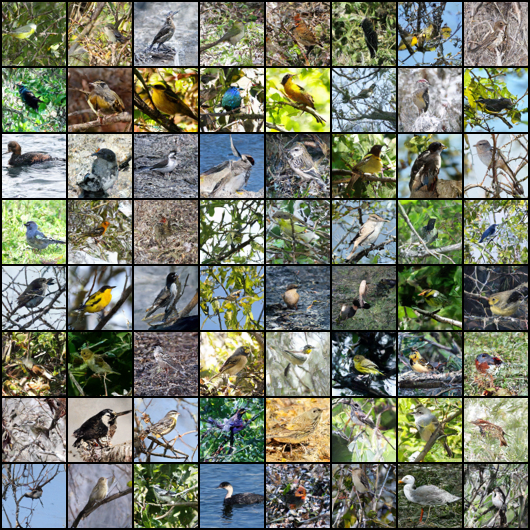}
   }
   \subfloat[\textbf{\textsf{\scriptsize{Generated Masks}}}] {
      \includegraphics[width = \columnwidth]{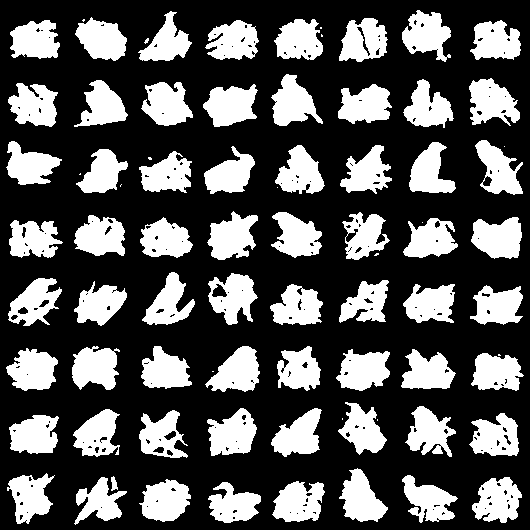}
   }
    \caption
      {%
        Generation on CUB.
        \label{fig:append:gen_cub}%
      }%
\end{figure*}

\begin{figure*}[htbp]
    \centering
   \subfloat[\textbf{\textsf{\scriptsize{Generated Images}}}] {
      \includegraphics[width = \columnwidth]{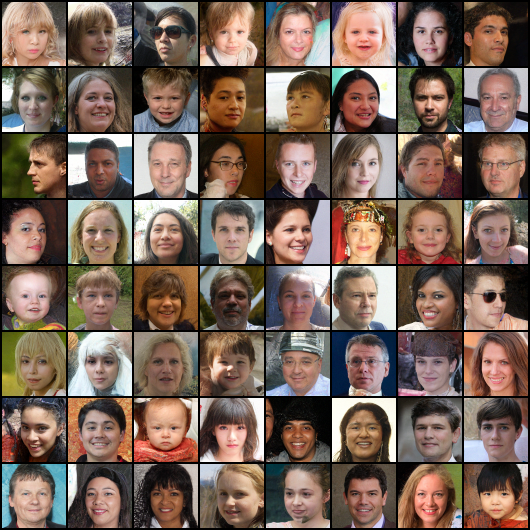}
   }
   \subfloat[\textbf{\textsf{\scriptsize{Generated Masks}}}] {
      \includegraphics[width = \columnwidth]{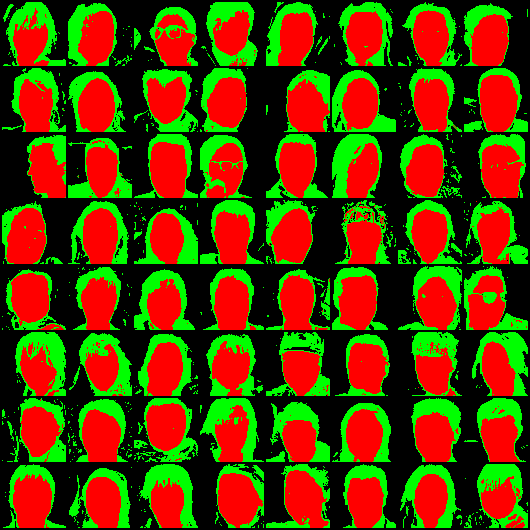}
   }
    \caption
      {%
        Generation on FFHQ.
        \label{fig:append:gen_ffhq}%
      }%
\end{figure*}

\begin{figure*}[htbp]
    \centering
   \subfloat[\textbf{\textsf{\scriptsize{Generated Images}}}] {
      \includegraphics[width = \columnwidth]{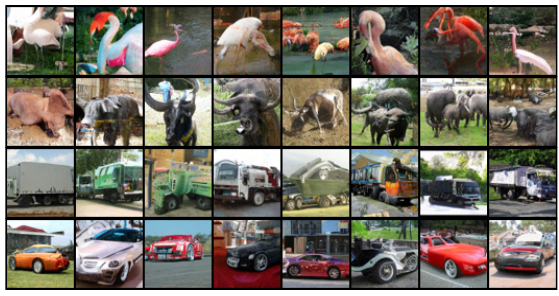}
   }
   \subfloat[\textbf{\textsf{\scriptsize{Generated Masks}}}] {
      \includegraphics[width = \columnwidth]{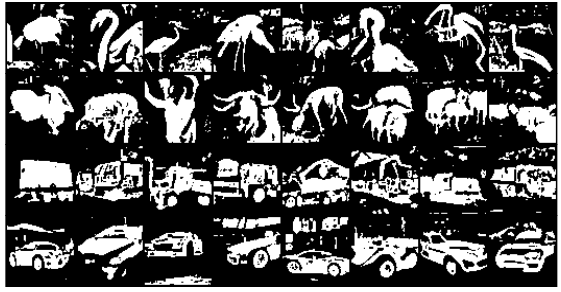}
   }
    \caption
      {%
        Conditional ImageNet generation.
        \label{fig:append:gen_imagenet}%
      }%
\end{figure*}

We show more generation results for Flower, CUB, FFHQ, and ImageNet (classes: flamingo, water buffalo, garbage truck, and sports car). As shown in Figures~\ref{fig:append:gen_flower},~\ref{fig:append:gen_cub},~\ref{fig:append:gen_ffhq}, and~\ref{fig:append:gen_imagenet}, our method consistently produces images and masks with high quality.
  \begin{table}[tbh]
  \setlength{\tabcolsep}{8pt}
    \begin{tabular}{lllcll}
    \toprule
    &\multicolumn{1}{c}{{ImageNet cls.}}
    &\multicolumn{1}{c}{Num of VOC val image} \\
    \midrule
    &1:aeroplane &895:warplane &136\\
    &2:bicycle &671:mountain-bike &108\\
    &3:bird &94:hummingbird &168\\
    &4:boat &814:speedboat &115\\
    &5:bottle &907:wine-bottle &133\\ 
    &6:bus &779:school-bus &114\\
    &7:car  &817:sports-car &191\\
    &8:cat &281:tabby &206\\
    &9:chair &765:rocking-chair &175\\ 
    &10:cow &346:water-buffalo &102\\
    &11:diningtable &532:dining-table &89\\
    &12:dog &153:maltese-dog &204\\
    &13:horse &603:horsecart &104\\ 
    &14:motorbike &670:motorscooter &117\\
    &15:person &981:ballplayer &584\\
    &16:potted plant &883:vase &116\\
    &17:sheep &348:ram &89\\
    &18:sofa &831:studio-couch &109\\
    &19:train &466:bullet-train &126\\
    &20:tv/monitor &664:monitor &106\\
    \bottomrule
    \end{tabular}
    \captionof{table}{{Class label mapping from ImageNet to VOC.}}
    \label{tab:cls_mapping}
    \end{table}
\subsection{Label Mapping for Zero-shot Transfer}\label{appendix:mapping}
At the current stage of diffusion model research, generation is not solved for PASCAL VOC when training from scratch without an extrernal large-scale dataset (\eg~LAION used in stable diffusion). There is no baseline DDPM model that can achieve this. As such, we adopt our conditional ImageNet model to perform zero-shot segmentation on VOC by mapping class labels from ImageNet to VOC.
Table~\ref{tab:cls_mapping} details the mapping rule.
  \begin{table}[tbh]
  \setlength{\tabcolsep}{15pt}
  \centering
    \begin{tabular}{llccc}
    \toprule
    &\multicolumn{1}{c}{VOC cls.}
    &\multicolumn{1}{c}{{Acc.}}
    &\multicolumn{1}{c}{mIOU} \\
    \midrule
    &1:aeroplane &0.82 &0.57\\
    &2:bicycle &0.79 &0.47\\
    &3:bird &0.83 &0.58\\
    &4:boat &0.81 &0.51\\
    &5:bottle &0.76 &0.47\\ 
    &6:bus &0.73 &0.54\\
    &7:car  &0.74 &0.48\\
    &8:cat &0.82 &0.66\\
    &9:chair &0.75 &0.64\\ 
    &10:cow &0.82 &0.45\\
    &11:diningtable &0.69 &0.62\\
    &12:dog &0.82 &0.67\\
    &13:horse &0.84 &0.53\\ 
    &14:motorbike &0.76 &0.52\\
    &15:person &0.77 &0.46\\
    &16:potted plant &0.74 &0.46\\
    &17:sheep &0.84 &0.64\\
    &18:sofa &0.73 &0.51\\
    &19:train &0.76 &0.56\\
    &20:tv/monitor &0.73 &0.47\\
    \midrule
    &average &0.78 &0.54 \\
    \bottomrule
    \end{tabular}
    \captionof{table}{{Accuracy and mIOU per class in VOC.}}
    \label{tab:cls_acc_voc}
    \end{table}

\subsection{Additional Zero-shot Results on VOC}\label{appendix:voc}
\begin{figure*}[htbp]
    \centering
   \subfloat[\textbf{\textsf{\scriptsize{Real Images}}}] {
      \includegraphics[width = \columnwidth]{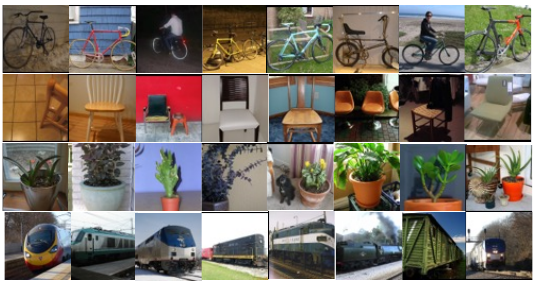}
   }
   \subfloat[\textbf{\textsf{\scriptsize{Segmentation}}}] {
      \includegraphics[width = \columnwidth]{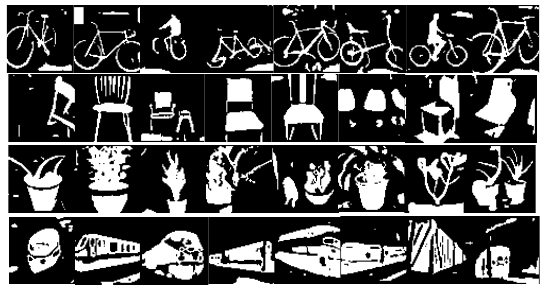}
   }
    \caption
      {%
        {Segmentation on VOC-2012.}
        \label{fig:append:seg_voc}%
      }%
  \end{figure*}

We report pixel accuracy and mIOU of each class in VOC in Table~\ref{tab:cls_acc_voc}, which demonstrates that our method can achieve reasonable high performance. We also provide more segmentation results of `bicycle', `chair', `potted plant' and `train' in Figure~\ref{fig:append:seg_voc}.

\subsection{Additional Zero-shot Results on DAVIS}\label{appendix:davis}
\begin{figure*}[tbh]
\begin{subfigure}[b]{\linewidth}
\centering
\includegraphics[width=\textwidth]{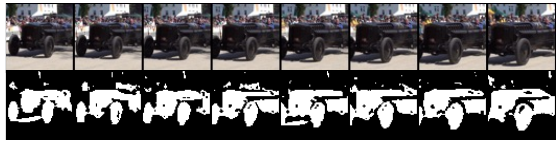}
\caption{\textbf{\textsf{\scriptsize{Frames of `Classic-car'}}}}
\label{seg:daivs_car}
\end{subfigure}
\begin{subfigure}[b]{\linewidth}
\centering
\includegraphics[width=\textwidth]{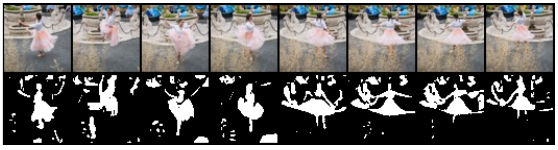}
\caption{\textbf{\textsf{\scriptsize{Frames of `Dance-jump'}}}}
\label{seg:daivs_dance}
\end{subfigure}
\vspace{-20pt}
\caption
  {%
    {Segmentation on DAVIS-2017.}
    \label{fig:append:seg_davis}%
  }%
\end{figure*}

We provide more DAVIS-2017 video segmentation results of `classic-car', `dance-jump' in Figure~\ref{fig:append:seg_davis}.

\end{document}